\documentclass[final]{article}

\PassOptionsToPackage{numbers, compress}{natbib}
\usepackage{neurips_2020}

\usepackage[utf8]{inputenc} 
\usepackage[T1]{fontenc}    
\usepackage{hyperref}       
\usepackage{url}            
\usepackage{booktabs}       
\usepackage{amsfonts}       
\usepackage{nicefrac}       
\usepackage{microtype}      

\usepackage[pdftex]{graphicx} 
\usepackage{subcaption}       
\usepackage{xcolor}           
\usepackage{amsmath}          
\usepackage{amssymb}          
\usepackage{wrapfig}          
\usepackage{lipsum}           
\usepackage{booktabs}         
\usepackage[noend]{algpseudocode}  

\newcommand{\note}[2]{}

\newcommand{\comment}[1]{}

\algnewcommand{\IIf}[1]{\State\algorithmicif\ #1\ \algorithmicthen}
\algnewcommand{\IElseIf}[1]{\State\algorithmicelse\ \algorithmicif\ 
  #1\ \algorithmicthen}
\algnewcommand{\IElse}{\State\algorithmicelse\ }
\algnewcommand{\EndIIf}{\unskip\ \algorithmicend\ \algorithmicif}

\newcommand{\indep}{\perp\!\!\!\!\perp}

\title{Algorithms for Causal Reasoning in Probability Trees}

\newcommand*\samethanks[1][\value{footnote}]{\footnotemark[#1]}

\author{%
  Tim Genewein\thanks{Equal contribution}, Tom McGrath\samethanks, Gr\'egoire Del\'etang\samethanks, Vladimir Mikulik\samethanks,\\
  \textbf{Miljan Martic, Shane Legg, Pedro A. Ortega}\thanks{Correspondence to \texttt{\{timgen|mcgrathtom|gdelt|vmikulik|pedroortega\}@google.com}}\\
  DeepMind\\London, UK
}

\begin{document}

\maketitle

\begin{abstract}
Probability trees are one of the simplest models of causal
generative processes. They possess clean semantics and---unlike
causal Bayesian networks---they can represent context-specific
causal dependencies, which are necessary for e.g.\ causal induction. 
Yet, they have received little attention
from the AI and ML community. Here we present concrete algorithms
for causal reasoning in discrete probability trees that cover the 
entire causal hierarchy (association, intervention, and 
counterfactuals), and operate on arbitrary propositional and causal
events. Our work expands the domain of causal reasoning to 
a very general class of discrete stochastic processes.
\end{abstract}

\section{Introduction}

\note{Pedro}{To write a note like this one, use \texttt{\char`\\note\{name\}\{text\}}.}
The formal treatment of causality is one of the major 
developments in AI and ML during the last two decades 
\citep{spirtes2000causation, pearl2009causality, dawid2015statistical, 
pearl2016causal, pearl2018book}. Causal reasoning techniques have
found their way into many ML applications~\citep{scholkopf2019causality},
and more recently also into RL~\citep{zinkevich2008regret,
ortega2010minimum, bareinboim2015bandits, lattimore2016causal,
dasgupta2019causal}, fairness~\citep{kusner2017counterfactual,
zhang2018fairness, chiappa2019path}, and
AI safety~\citep{everitt2019understanding, carey2020incentives}, 
to mention some. This broad adoption rests
on the greater explanatory power 
delivered by causal models than by those based on statistical
dependence alone~\citep{pearl2018book, arjovsky2019invariant}.

Causal dependencies can be described using the language
of \emph{causal Bayesian networks} (CBNs) and \emph{structural 
causal models},
which elegantly tie together graphical properties with conditional 
independence relations and effects of interventions
\citep{spirtes2000causation, pearl2009causality}. 
While very versatile, CBNs have
well-known limitations: the collection of conditional 
independence properties they can express are very 
specific~\citep{dawid2010beware}, requiring 
the causal relations among random variables
to form a partial order.

\begin{figure}[h]
    \centering
    \includegraphics[width=\textwidth]{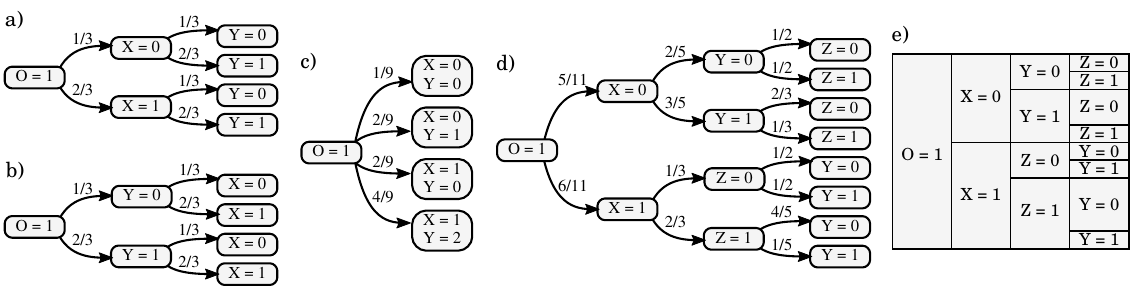}
    \caption{Probability trees. Panels (a), (b), and (c) show the same
    joint distribution over $X$ and $Y$. They differ in that (a) assumes
    $X \rightarrow Y$, (b) assumes $Y \rightarrow X$, whereas (c) 
    does not assume a causal dependency ($X \not\rightarrow Y \wedge
    Y \not\rightarrow X$). Panel~(d) will be our running example
    throughout the paper. It is a probability tree where
    $Y \rightarrow Z$ when $X = 0$ and $Z \rightarrow Y$ when $X = 1$.
    Such conditional causal dependencies cannot be expressed using
    a CBN. Panel (e) shows a probability tree
    mass diagram, which is an alternative representation of the 
    probability tree (d) that emphasizes 
    the probability mass of events (encoded as the height of a box).
    By convention we bind~$O=1$ (as in omega ``$\Omega$''
    for a sample space) at the root node.}
    \label{fig:intro-examples}
\end{figure}

Here we work with an alternative representation: \emph{discrete probability trees}, sometimes also called \emph{staged tree models}
\citep{shafer1996art, ortega2011bayesian, gorgen2017algebraic,
gorgen2018equivalence}.
A probability tree  is one of the simplest models 
for representing the causal generative process of a random experiment 
or stochastic process (Figure~\ref{fig:intro-examples}). 
The semantics are self-explanatory:
each node in the tree corresponds to a potential state
of the process, and the arrows indicate both the 
probabilistic transitions and the causal dependencies 
between them. Unlike CBNs, probability trees can model
context-specific causal dependencies (see e.g.\ 
Figure~\ref{fig:intro-examples}c).
However, probability trees do not explicitly represent
conditional independencies, and thus, when a distribution and
its causal relations admit a representation both as a probability
tree and a CBN, the latter is more compact. 

In spite of their simplicity and expressiveness, 
probability trees have not enjoyed the same attention 
as CBNs and structural causal models in the statistical and 
machine learning literature. In this work,
we attempt to remedy this through the following contributions.
Focusing on finite probability trees, our work is the first
to provide concrete algorithms for~(a) computing minimal 
representations of \emph{arbitrary events} formed through 
propositional calculus and causal precedences; and~(b) computing 
the \emph{three fundamental operations} of the causal 
hierarchy~\citep{pearl2009causality}, namely conditions, 
interventions, and counterfactuals. We also provide an
interactive tutorial available online\footnote{%
\url{https://github.com/deepmind/deepmind-research/tree/master/causal_reasoning}}, 
containing many concrete examples.

\paragraph{Formal definition.}
While probability trees can be abstractly axiomatized or
described as colored graphs \citep{shafer1996art, ortega2015subjectivity, 
gorgen2017algebraic}, here we adopt a recursive definition 
that is closer to a computational implementation. 
We define a \emph{node} $n \in \mathcal{N}$ in the tree as a tuple 
$n = (u, \mathcal{S}, \mathcal{C})$ where: $u \in \mathbb{N}$
is a unique numerical identifier for the node within the tree; 
$\mathcal{S}$ is a list of statements such as 
`$X = 0$' and `$W = \text{rainy}$'; and
$\mathcal{C}$ is a (possibly-empty) ordered set of transitions 
$(p, m) \in [0, 1] \times \mathcal{N}$ where $p$ is the
transition probability to the child node $m$. We will represent statements
such as `$X = 1$' as a tuple~$(X, 1) \in \mathcal{X} \times \mathcal{V}_X$ 
where $\mathcal{X}$ is the set of variables and $\mathcal{V}_X$
is the range of the variable~$X$. Obviously,
the transition probabilities must sum up to one. The root
is the unique node with no parents, and a leaf is a node with 
an empty set of transitions. A \emph{(total) realization} in the 
probability tree is a path from the root to a leaf, and its probability
is obtained by multiplying the transition probabilities along the path; 
and a \emph{partial realization} is any connected sub-path within a 
total realization. When entering a node, the process binds the 
listed random variables to definite values.

\section{Events}

We have seen how to represent the realizations and causal 
dependencies of a random experiment using a probability tree. 
Here we show how to represent and calculate events.

An \emph{event} is a collection of total realizations. We can 
describe events using using propositions about random 
variables (e.g.\ `$X=0$',  `$Y=1$'). For instance, 
the event `$X = 0$' is the set of all total realizations
that traverse a node with the statement `$X = 0$'.
Furthermore, we can use logical connectives
of negation (\textsc{Not}, $\neg$), conjunction (\textsc{And}, $\wedge$), 
and disjunction (\textsc{Or}, $\vee$) to state composite events,
such as `$\neg(X=0 \wedge Y=1)$'. In addition, we can also
use precedence (\textsc{Prec}, $\prec$) for describing events
that meet a causal condition.

\subsection{Min-cuts}\label{sec:mincut}

We can represent events using \emph{cuts}---a collection of nodes
with probabilities summing up to one that are mutually exclusive 
\citep{shafer1996art}. In particular, we will focus our attention on
\emph{min-cuts} (minimal cuts). 
A min-cut is a minimal representation of an event in terms of 
the nodes of a probability tree. The min-cut of an event collects
the smallest number of nodes in the probability tree that resolves
whether an event has occurred or not. In other words, if a realization
hits a node in the min-cut, then we know for sure whether the event 
has occurred\footnote{In measure theory, a similar notion to a 
min-cut would be the smallest algebra that renders an event measurable.}.

\begin{figure}[t]
    \centering
    \includegraphics[width=\textwidth]{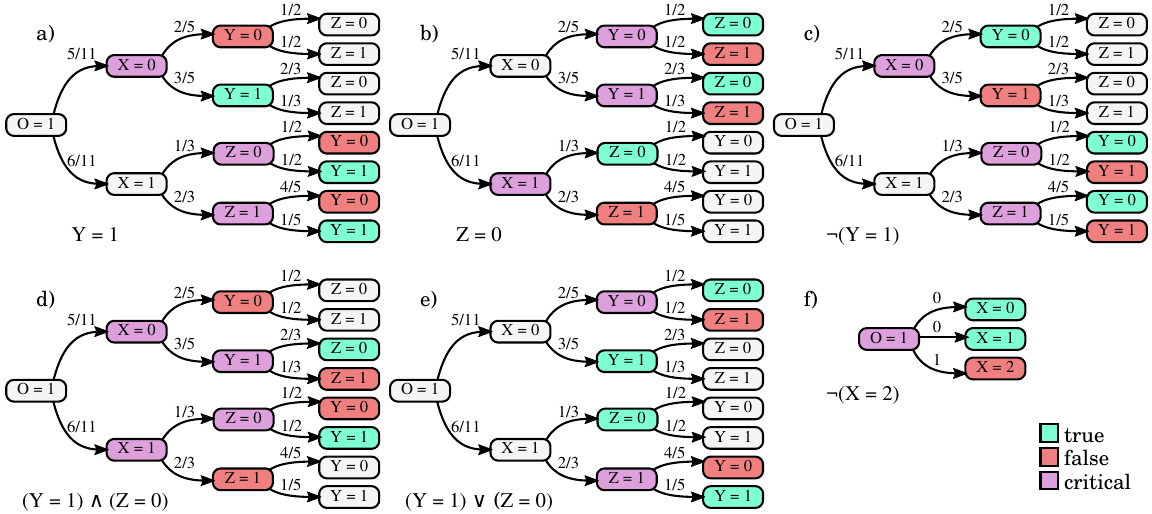}
    \caption{Min-cuts and critical sets. Panels (a)--(e) are
    min-cuts (red and green) and critical sets (purple) for different
    events within the probability tree of
    Figure~\ref{fig:intro-examples}d. They correspond to:
    (a) $Y=1$; (b) $Z=0$; (c) the negation $\neg(Y=1)$; (d) 
    the conjunction $Y=1 \wedge Z=0$; and (e) the disjunction 
    $Y=1 \vee Z=0$. Negations swap true and false sets; 
    conjunctions pick, for each realization, the first and last
    occurrence of a false and true node respectively; and
    disjunctions invert the selection order of conjunctions.
    Panel (f) shows the min-cut for the
    event $\neg(X=2)$. Notice that the true set contains both
    $X=0$ and $X=1$, which can never occur probabilistically.
    This example illustrates that min-cuts are determined by 
    the logical truth and not by the probabilistic truth of
    events. }
    \label{fig:mincut_critical}
\end{figure}

Our custom notion of min-cuts furthermore distinguishes between the
nodes that render the event true from the nodes that render the 
event false. More precisely, we describe an event using a cut 
$\delta = (\mathcal{T}, \mathcal{F})$, where the true set $\mathcal{T}$
and the false set $\mathcal{F}$ contain all the nodes where the event
becomes true or false respectively. Figure~\ref{fig:mincut_critical}
shows examples of min-cuts for \emph{simple events} (defined here as events
that bind a single random variable to a constant), negation,
conjunction, and disjunction.

The (mostly recursive) pseudo-algorithms for computing the min-cuts are 
listed in Figure~\ref{alg:mincut}.
They compute the min-cuts for simple events (\textsc{Prop}),
negations (\textsc{Neg}),
conjunctions (\textsc{And}), and disjunctions (\textsc{Or}), returning a
pair $\delta = (\mathcal{T}, \mathcal{F})$ of a true and a false set
of unique node identifiers. These algorithms assume that every simple
event (e.g.\ $X = 1$) is \emph{well-formed}, that is, 
every total realization must contain
one and only one node within its path where the simple event is either
true ($X = 1$) or false ($X \not= 1$). Or, in other words, every random
variable must be bound to a unique value by the time the stochastic process 
terminates at a leaf. If simple events are well-defined, then so are their
compositions through the Boolean operators.

It is worth pointing out that we distinguish between probabilistic and
logical truth. A min-cut for a given event is determined solely through
the skeleton of the probability tree without taking into account the
transition probabilities (see the example in Figure~\ref{fig:mincut_critical}f).

Once a min-cut for an event has been determined, then the probability of
said event is given by the sum of the probability of the true nodes.
Probability distributions, expectations, etc.\ are then determined
from these probabilities in the obvious manner.

\subsection{Mechanisms and critical sets}

For every event, we can also identify the nodes where the very
next transition determines whether a given event will not occur. 
These are the \emph{critical nodes}. Formally,
given the min-cut for an event, we define the \emph{critical set}
as the set of all nodes that are parents of a node in the false set.
They are highlighted in purple in Figure~\ref{fig:mincut_critical}.
The critical set can be thought of as the 
collection of mechanisms in the tree that we need to manipulate 
in order to bring about a given event. Critical nodes are the 
\emph{Markov blankets}: all the variables bound within a path from the root
to the critical node constitute the ``exogenous variables''
for the mechanisms downstream. They are (implicitly) used in defining the
three operations of the causal hierarchy \citep{pearl2009causality}
presented later.

\newpage
\begin{figure}[t]
\noindent\fbox{
\begin{minipage}{0.48\textwidth}
    \footnotesize
 
    \begin{algorithmic}[1]
    \Function{Prop}{$n, s$}
        \State $\rhd$ $n$ : Node, $s$ : Statement
        \State Let $n = (u, \cal{S}, \cal{C})$ and $s = (X_s, V_s)$
        \State $\rhd$ Base case, find $s$ within node:
        \ForAll{$(X, V) \in \cal{S}$}
            \If{$X_s = X$}
                \If{$V_s = X$} 
                    \State\Return $(\{u\}, \varnothing)$
                \EndIf
            \IElse \Return $(\varnothing, \{u\})$
            \EndIf
        \EndFor
    \If{$\cal{C} = \varnothing$}
        \State \textbf{error:} $s$ cannot be resolved.
    \EndIf
  
    \State $\rhd$ Statement not found, recurse:
    \State $\delta \leftarrow (\varnothing, \varnothing)$
    \ForAll{$c = (p_c, n_c) \in \cal{C}$}
        \State $\delta_c \leftarrow\ $\textsc{Prop}$(n_c, s)$
        \State Let $\delta_c = ({\cal T}_c, {\cal F}_c)$
            and $\delta = (\cal{T}, \cal{F})$
        \State $\delta \leftarrow (\mathcal{T} \cup \mathcal{T}_c, 
            \mathcal{F} \cup \mathcal{F}_c)$
    \EndFor
    \State $\rhd$ Consolidate:
    \IIf{$\mathcal{F} = \varnothing$}
        $\delta \leftarrow (\{u\}, \varnothing)$
    \IElseIf{$\mathcal{T} = \varnothing$}
        $\delta \leftarrow (\varnothing, \{u\})$
  
    \State \Return $\delta$
    \EndFunction
    \end{algorithmic}

    \begin{algorithmic}[1]
    \Function{Neg}{$\delta$: Min-Cut}
        \State $\rhd$ Switch true and false:
        \State Let $\delta = (\mathcal{T}, \mathcal{F})$
        \State \Return $(\mathcal{F}, \mathcal{T})$
    \EndFunction
    \end{algorithmic}
\end{minipage}
\hfill 
\begin{minipage}{0.48\textwidth}
    \footnotesize

    \begin{algorithmic}[1]
    \Function{And}{$n, \delta_1, \delta_2$}
        \State $\rhd$ $n$: Node; $\delta_1, \delta_2$: Min-Cut
        \State \Return \textsc{And-R}$(n, \delta_1, \delta_2, \bot, \bot)$
    \EndFunction

    \Function{And-R}{$n, \delta_1, \delta_2, e_1, e_2$}
        \State $\rhd$ $n$: Node; $\delta_1$, $\delta_2$: Min-Cut;
        \State \phantom{$\rhd$} $e_1, e_2$: Boolean
        \State Let $n = (u, \mathcal{S}, \mathcal{C})$
        \State Let $\delta_1 = (\mathcal{T}_1, \mathcal{F}_1)$
            and $\delta_2 = (\mathcal{T}_2, \mathcal{F}_2)$
        \State$\rhd$ Base case:
        \IIf{$u \in \mathcal{F}_1 \cup \mathcal{F}_2$}
            \Return $(\varnothing, \{u\})$
        \IIf{$u \in \mathcal{T}_1$} $e_1 \leftarrow \top$
        \IIf{$u \in \mathcal{T}_2$} $e_2 \leftarrow \top$
        \IIf{$e_1 \wedge e_2$} \Return $(\{u\}, \varnothing)$
  
        \State $\rhd$ Recurse:
        \State $\delta \leftarrow (\varnothing, \varnothing)$
        \ForAll{$(p_c, n_c) \in \mathcal{C}$}
            \State Let $\delta = (\mathcal{T}, \mathcal{F})$
            \State \mbox{$(\mathcal{T}_c, \mathcal{F}_c) 
                \leftarrow $\ \textsc{And-R}$(n_c, \delta_1, \delta_2, e_1, e_2)$}
            \State $\delta \leftarrow 
                (\mathcal{T} \cup \mathcal{T}_c, \mathcal{F} \cup \mathcal{F}_c)$
        \EndFor
        \State $\rhd$ Consolidate:
        \IIf{$\mathcal{F} = \varnothing$}
            $\delta \leftarrow (\{u\}, \varnothing)$
        \IElseIf{$\mathcal{T} = \varnothing$}
            $\delta \leftarrow (\varnothing, \{u\})$
        \State \Return $\delta$
    \EndFunction
    \end{algorithmic}

    \begin{algorithmic}[1]
    \Function{Or}{$n$: Node, $\delta_1, \delta_2$: Min-Cut}
        \State $\rhd$ Use De Morgan's rule:
        \State \mbox{\Return \textsc{Neg}$($\textsc{And}$(n,$\ \textsc{Neg}$(\delta_1),$
        \textsc{Neg}$(\delta_2)))$}
    \EndFunction
    \end{algorithmic}
\end{minipage}
}
\caption{Algorithms for determining min-cuts. All functions calls are 
\emph{pass-by-value} (that is, deep copies of their arguments). The recursive function
\textsc{Prop}$(n, s)$ takes a root node $n$ and a single statement $s=(X_s,V_s)$
and returns the min-cut $\delta$ for the event. \textsc{Neg}$(\delta)$ computes
the min-cut for the negation of $\delta$. \textsc{And}$(n, \delta_1, \delta_2)$
computes the conjunction of the min-cuts $\delta_1$ and $\delta_2$ for a probability
tree with root $n$ using the auxiliary function \textsc{And-R}. Finally, \textsc{Or}
computes disjunctions.}\label{alg:mincut}
\end{figure}

\subsection{Min-cuts for causal events}\label{sec:mincut-causal}
\begin{wrapfigure}{r}{0.45\textwidth}
\vspace{-32.5pt}
\noindent\fbox{
\begin{minipage}{0.42\textwidth}
    \footnotesize
    \begin{algorithmic}[1]
    \Function{Prec}{$n, \delta_c, \delta_f$}
        \State $\rhd$ $n$: Node; $\delta_c, \delta_e$: Min-Cut
        \State \Return \textsc{Prec-R}$(n, \delta_c, \delta_e, \bot)$
    \EndFunction

    \Function{Prec-R}{$n, \delta_c, \delta_e, f$}
        \State $\rhd$ $n$: Node; $\delta_c$, $\delta_e$: Min-Cut;
        \State \phantom{$\rhd$} $f$: Boolean
        \State Let $n = (u, \mathcal{S}, \mathcal{C})$
        \State Let $\delta_c = (\mathcal{T}_c, \mathcal{F}_c)$
            and $\delta_e = (\mathcal{T}_e, \mathcal{F}_e)$
            
        \State$\rhd$ Base case:
        \If{$f = \bot$}
            \If{$u \in \mathcal{T}_e \cup \mathcal{F}_e \cup \mathcal{F}_c$}
                \State \Return $(\varnothing, \{u\})$
            \EndIf
            \IIf{$u \in \mathcal{T}_c$}
                $f \leftarrow \top$
        \ElsIf{$f = \top$}
            \IIf{$u \in \mathcal{T}_c$}
                \Return $(\{u\}, \varnothing)$
            \IIf{$u \in \mathcal{F}_c$}
                \Return $(\varnothing, \{u\})$
        \EndIf

        \State $\rhd$ Recurse:
        \State $\delta \leftarrow (\varnothing, \varnothing)$
        \ForAll{$(p_c, n_c) \in \mathcal{C}$}
            \State Let $\delta = (\mathcal{T}, \mathcal{F})$
            \State \mbox{$(\mathcal{T}_c, \mathcal{F}_c) 
                \leftarrow $\ \textsc{Prec-R}$(n_c, \delta_c, \delta_e, f)$}
            \State $\delta \leftarrow 
                (\mathcal{T} \cup \mathcal{T}_c, \mathcal{F} \cup \mathcal{F}_c)$
        \EndFor
        \State $\rhd$ Consolidate:
        \IIf{$\mathcal{F} = \varnothing$}
            $\delta \leftarrow (\{u\}, \varnothing)$
        \IElseIf{$\mathcal{T} = \varnothing$}
            $\delta \leftarrow (\varnothing, \{u\})$
        \State \Return $\delta$
    \EndFunction
    \end{algorithmic}
\end{minipage}
}
\caption{Precedence min-cut  algorithm}\label{alg:preceeds}
\vspace{-150pt}
\end{wrapfigure}

\begin{figure}[h]
\begin{minipage}{0.5\textwidth}
    \centering
    \includegraphics[width=0.8\textwidth]{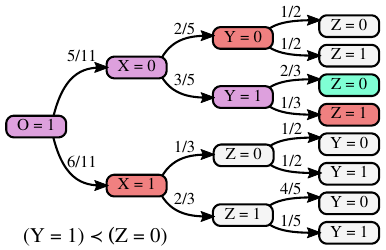}
    \caption{Min-cut for a causal event}
    \label{fig:prec}
\end{minipage}
\end{figure}

A statement such as ``the event where $Y = 1$ precedes $Z = 0$'',
written $Y = 1 \prec Z = 0$, cannot be stated logically. Rather,
it is a \emph{causal statement} that requires the causal 
relations provided in the probability tree. Figure~\ref{fig:prec}
illustrates a min-cut for a precedence event. This relation
can be combined arbitrarily with the logical connectives
to form composite events.
Precedences can be computed recursively as shown in the
pseudo-code in Figure~\ref{alg:preceeds}. The function 
\textsc{Prec}$(n, \delta_c, \delta_e)$ takes a root~$n$ 
and two min-cuts $\delta_c$ and $\delta_e$, one for the cause
and the effect respectively, and returns the min-cut for
the event where the precedence relation holds.

\newpage
\section{Conditions}

\begin{figure}[t]
    \centering
    \includegraphics[width=\textwidth]{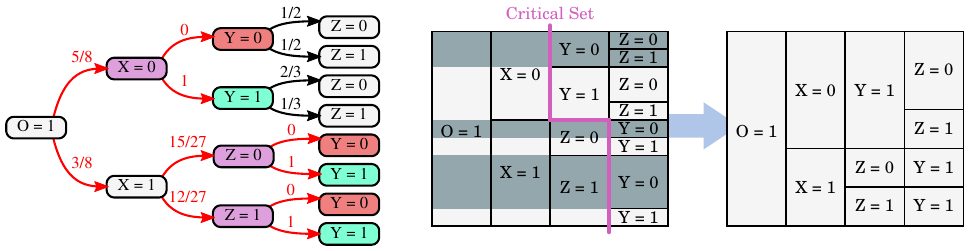}
    \caption{Seeing $Y=1$. Conditioning on an event proceeds in two steps:
        first, we remove the probability mass of the realizations passing
        through the false set of the event's min-cut; then we renormalize
        the probabilities. We can do this recursively by aggregating
        the original probabilities of the true set. The left panel shows
        the result of conditioning the probability tree in 
        Figure~\ref{fig:intro-examples}d on the event $Y=1$, which also
        highlights the modified transition probabilities in red.
        The right panel shows the same operation in a probability
        mass diagram.}
    \label{fig:see}
\end{figure}

\begin{wrapfigure}{r}{0.43\textwidth}
\vspace{-10pt}
\noindent\fbox{
\begin{minipage}{0.40\textwidth}
    \footnotesize
    \begin{algorithmic}[1]
    \Function{See}{$n$: Node, $\delta$: Min-Cut}
        \State $(n, l, p) \leftarrow$\ \textsc{See-R}$(n, \delta, 1)$
        \State \Return $n$
    \EndFunction

    \Function{See-R}{$n, \delta, q$}
        \State $\rhd$ $n$: Node; $\delta$: Min-Cut; $q$: [0, 1]
        \State Let $n = (u, \mathcal{S}, \mathcal{C})$ 
            and $\delta = (\mathcal{T}, \mathcal{F})$
        \State $\rhd$ Base case:
        \If{$u \in \mathcal{T}$} 
            \State \Return $(n, 1, p)$
        \EndIf
        \If{$u \in \mathcal{F}$}
            \State \Return $(n, 0, 0)$
        \EndIf
        \State $\rhd$ Recurse:
        \State $\mathcal{D} \leftarrow \varnothing$
        \State $\sigma_l \leftarrow 0$ and $\sigma_p \leftarrow 0$
        \ForAll{$(p, \tilde{n}) \in \mathcal{C}$}
            \State $(\tilde{n}, l, p) 
                \leftarrow$\ \textsc{See-R}$(\tilde{n}, \delta, q \cdot p)$
            \State $\mathcal{D} \leftarrow \mathcal{D} \cup \{(\tilde{n}, l, p)\}$
            \State $\sigma_l \leftarrow \sigma_l + l$
            \State $\sigma_p \leftarrow \sigma_p + p$
        \EndFor
        \State $\rhd$ Normalize:
        \State $\mathcal{C} \leftarrow$\ 
            \textsc{Normalize}$(\mathcal{D}, \sigma_l, \sigma_p)$
        \State $n \leftarrow (u, \mathcal{S}, \mathcal{C})$
        \State \Return $(n, 1, \sigma_p)$
    \EndFunction
    \end{algorithmic}
    \bigskip
    \begin{algorithmic}[1]
    \Function{Normalize}{$\mathcal{D}, \sigma_l, \sigma_p$}
        \If{$\sigma_p > 0$}
            \State $\mathcal{C} \leftarrow 
                \bigl\{ \bigl( \frac{p}{\sigma_p}, n \bigr)
                \bigl| (n, l, p) \in \mathcal{D} \bigr\}$
        \Else
            \State $\mathcal{C} \leftarrow 
                \bigl\{ \bigl( \frac{l}{\sigma_l}, n \bigr)
                \bigl| (n, l, p) \in \mathcal{D} \bigr\}$
        \EndIf
        \State \Return $\mathcal{C}$
    \EndFunction
    \end{algorithmic}
\end{minipage}
}
\caption{Conditioning algorithm}\label{alg:conditioning}
\vspace{-6pt}
\end{wrapfigure}

In a probability tree, conditioning is the act of updating the
transition probabilities after an event is revealed to be true. 
This operation allows answering questions of the form
\[
    P(A \mid B)
\]
that is, ``what is the probability of the event~$A$ given that 
the event~$B$ is true?'' Here, $A$ could e.g.\ be an event
occurring downstream (prediction) or upstream
(inference) of the event~$B$. The operation is illustrated in
Figure~\ref{fig:see}.

Computationally, conditioning acts as a filter: given an
event~$B$, it removes all the probability mass from the
total realizations that are within the event $\neg B$,
renormalizing the remaining probability mass globally.
This operation can be performed recursively. Given
a probability tree and an event's min-cut, we recursively
descend the tree depth-first until reaching a node within 
the min-cut. If the node is within the true set, we compute
its probability and return it; otherwise we return zero.
Then, during the recursive ascent, we equate the transition probabilities
with returned probabilities normalized by their aggregate
sum. This sum is then returned upstream and the process is repeated.

\begin{wrapfigure}{r}{0.43\textwidth}
\vspace{-27pt}
    \centering
    \includegraphics[width=0.40\textwidth]{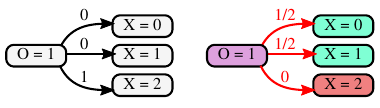}
    \caption{Conditioning on $\neg(X=2)$.}
    \label{fig:cond-int-zero}
\vspace{-10pt}
\end{wrapfigure}

A special case occurs when we condition on an event with
probability zero, or more generally, when there exists
a node where every next transition leading into the true
set has probability zero. Such transitions cannot be normalized
by dividing by a normalizing constant. In this case, measure theory
requires us to choose a \emph{version} of the conditional probability.
Many choices exist, but for concreteness, here we have adopted a uniform
distribution (see Figure~\ref{fig:cond-int-zero}).

The algorithm for conditioning is shown in Figure~\ref{alg:conditioning}.
\textsc{See}$(n, \delta)$ takes a root~$n$ of a probability tree and a min-cut~$\delta$
and returns the root of a new, conditioned probability tree. The logical
structure of the original probability tree is preserved; only the transition
probabilities upstream of the min-cut change. Finally, conditioning is
a commutative operation.

\newpage
\section{Interventions}

\begin{figure}[t]
    \centering
    \includegraphics[width=\textwidth]{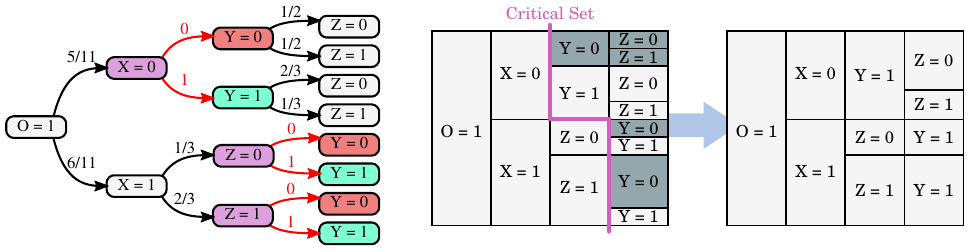}
    \caption{Doing $Y \leftarrow 1 := \text{do}(Y = 1)$. 
        An intervention proceeds in two steps:
        first, it selects the partial realizations starting in a critical
        node and ending in a leaf that traverse the false set of the
        event's min-cut; then it removes their probability mass,
        renormalizing the probabilities leaving the critical set.
        The left panel shows the result of intervening the probability
        tree from Figure~\ref{fig:intro-examples}d with $Y \leftarrow 1$.
        The right panel illustrates the procedure on the corresponding
        probability mass diagram.}
    \label{fig:do}
\end{figure}

\begin{wrapfigure}{r}{0.43\textwidth}
\vspace{-12pt}
\noindent\fbox{
\begin{minipage}{0.40\textwidth}
    \footnotesize
    \begin{algorithmic}[1]
    \Function{Do}{$n$: Node, $\delta$: Min-Cut}
        \State $(n, l) \leftarrow$\ \textsc{Do-R}$(n, \delta)$
        \State \Return $n$
    \EndFunction

    \Function{Do-R}{$n$: Node, $\delta$: Min-Cut}
        \State Let $n = (u, \mathcal{S}, \mathcal{C})$ 
            and $\delta = (\mathcal{T}, \mathcal{F})$
        \State $\rhd$ Base case:
        \IIf{$u \in \mathcal{T}$} \Return $(n, \top)$
        \IIf{$u \in \mathcal{F}$} \Return $(n, \bot)$
        \State $\rhd$ Recurse:
        \State $\mathcal{D} \leftarrow \varnothing$
        \State $\sigma_l \leftarrow 0$ and $\sigma_p \leftarrow 0$
        \ForAll{$(p, \tilde{n}) \in \mathcal{C}$}
            \State $(\tilde{n}, \beta) \leftarrow$\ 
                \textsc{Do-R}$(\tilde{n}, \delta)$
            \If{$\beta = \top$}
                \State $\mathcal{D} \leftarrow 
                    \mathcal{D} \cup \{(\tilde{n}, 1, p)\}$
                \State $\sigma_l \leftarrow \sigma_l + 1$
                \State $\sigma_p \leftarrow \sigma_p + p$
            \Else
            \State $\mathcal{D} \leftarrow 
                \mathcal{D} \cup \{(\tilde{n}, 0, p)\}$
            \EndIf
        \EndFor
        \State $\rhd$ Normalize:
        \State $\mathcal{C} \leftarrow$\ 
            \textsc{Normalize}$(\mathcal{D}, \sigma_l, \sigma_p)$
        \State $n \leftarrow (u, \mathcal{S}, \mathcal{C})$
        \State \Return $(n, \top)$
    \EndFunction
    \end{algorithmic}
\end{minipage}
}
\caption{Intervention algorithm}\label{alg:intervening}
\vspace{-10pt}
\end{wrapfigure}

In a probability tree, intervening is the act of minimally
changing the transition probabilities in order to bring about
a desired event with probability one. This operation allows 
answering questions of the form
\[
    P(A \mid \text{do}(B))
\]
that is, ``what is the probability of the event~$A$ given that 
the event~$B$ was made true?'' The operation is illustrated in
Figure~\ref{fig:see}. The crucial difference to conditions
is that interventions only affect realizations downstream 
of the critical set, whereas conditions also provide information
on what happened upstream of the set.

Computationally, interventions are simpler than conditions.
We compute interventions using a depth-first search recursive
algorithm, which descents down the probability tree until
the min-cut of the desired event is reached. Then, we
remove the probability mass of the transitions leading
into the nodes of the false set and renormalize the transition
probabilities into the nodes of the true set. This normalization
is local, relative to the subtree rooted at the 
parent critical node. As in conditioning,
intervening on events having nodes with probability zero in
the false set requires special treatment, and we settle
for the same solution as before.

The pseudo-algorithm for computing interventions is shown in
Figure~\ref{alg:intervening}. The function \textsc{Do}$(n, \delta)$ takes
a root $n$ and a min-cut $\delta$, and returns the root of a
new, intervened probability tree. 

Interventions do not alter the structure of the original 
probability tree; they only change the transition probabilities 
emanating from the critical set. Furthermore, they are
commutative, as is conditioning. However, conditions and
interventions do \emph{not} commute. For instance,
for the probability tree in Figure~\ref{fig:intro-examples}d,
\[
  P(X = 0 \mid Y \leftarrow 1; Z = 0) = \frac{5}{8} \not=
  P(X = 0 \mid Z = 0; Y \leftarrow 1) = \frac{3}{5}
\]
where the semicolon ``$;$'' denotes the sequential composition 
of the operators. 

It should be stressed that interventions on events are strictly
more general than interventions on random variables. Thus, 
unlike in CBNs, an intervention does not necessarily assign
a unique value to a manipulated random variable. 
Rather, depending on the critical
set, the intervention could assign different values to the
same random variables in separate branches of the tree, 
or in fact even manipulate different random variables in
every branch (a context-dependent ``recipe''), in order to 
bring about a desired effect.

\newpage
\section{Counterfactuals}

\begin{figure}[t]
    \centering
    \includegraphics[width=0.9\textwidth]{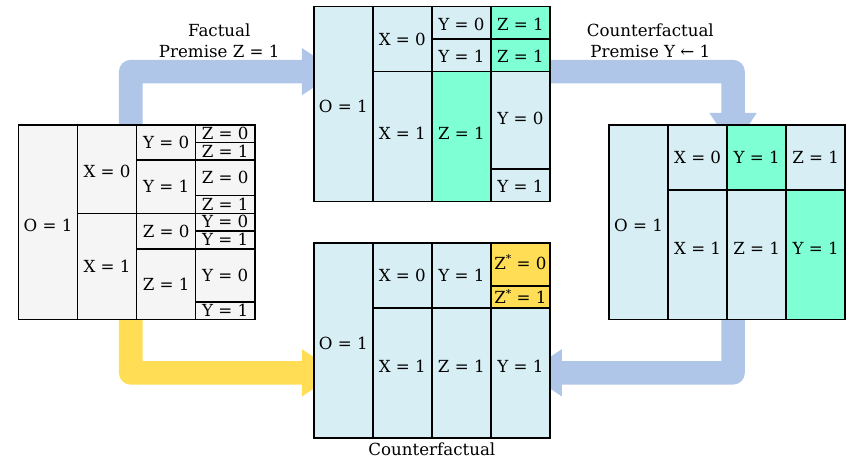}
    \caption{The counterfactual probability tree generated by 
        imposing $Y \leftarrow 1$, given the factual premise $Z = 1$.
        Starting from a reference probability tree describing the original state of uncertainty, we perform an update to reflect the current state of affairs (``see $Z=1$''), then modify it using a counterfactual premise (``do $Y \leftarrow 1$''). To form the counterfactual, we then restore the events downstream of the counterfactual premise's min-cut to their original state of uncertainty. These events then span a new scope with copies of the original random variables (marked with ``$\ast$''), ready to adopt new values. In particular note that $Z^\ast=0$ can happen in our alternative reality, even though we know that $Z=1$.}
    \label{fig:cf}
\end{figure}

\begin{wrapfigure}{r}{0.43\textwidth}
\vspace{-12pt}
\noindent\fbox{
\begin{minipage}{0.40\textwidth}
    \footnotesize
    \begin{algorithmic}[1]
    \Function{CF}{$n, m$: Node, $\delta$: Min-Cut}
        \State $m \leftarrow$\ \textsc{Do}$(m, \delta)$
        \State $(n, l) \leftarrow$\ \textsc{CF-R}$(n, m, \delta)$
        \State \Return $n$
    \EndFunction

    \Function{CF-R}{$n, m, \delta$}
        \State $\rhd$ $n, m$: Node; $\delta$: Min-Cut
        \State Let $n = (u, \mathcal{S}, \mathcal{C}_n)$
        \State Let $m = (u, \mathcal{S}, \mathcal{C}_m)$
        \State Let $\delta = (\mathcal{T}, \mathcal{F})$
        \State $\rhd$ Base case:
        \IIf{$u \in \mathcal{T} \cup \mathcal{F}$} 
          \Return $n$
        \State $\rhd$ Recurse:
        \State $\mathcal{C} \leftarrow \varnothing$
        \ForAll{$(c, d) \in$\ \textsc{Zip}$(\mathcal{C}_n, \mathcal{C}_m)$}
            \State Let $c = (p, \tilde{n})$, $d = (q, \tilde{m})$
            \State $\tilde{n} 
                \leftarrow$\ \textsc{CF-R}$(\tilde{n}, \tilde{m}, \delta)$
            \State $\mathcal{C} \leftarrow \mathcal{C} \cup \{(q, \tilde{n})\}$
        \EndFor
        \State $\rhd$ Update probabilities of children:
        \State $n \leftarrow (u, \mathcal{S}, \mathcal{C})$
        \State \Return $n$
    \EndFunction
    \end{algorithmic}
\end{minipage}
}
\caption{Counterfactual algorithm}\label{alg:counterfactual}
\vspace{-10pt}
\end{wrapfigure}

In a probability tree, a counterfactual is a statement about a
subjunctive (i.e.\ possible or imagined) event that could have happened
had the stochastic process taken a different course during its realization.
This operation allows evaluating probabilities of the form
\[
    P(A_C \mid B)
\]
that is, ``Given that $B$ is true, what would the probability 
of $A$ be if $C$ \emph{were} true?''. Here, $A_C$ denotes the 
subjunctive event $A$ under the counterfactual assumption that
the event~$C$ has occurred (i.e.\ a potential response), 
and~$B$ is the indicative (i.e.\ factual) assumption. 

A counterfactual probability tree is obtained by modifying
a reference probability tree through factual statements
(i.e.\ conditioning and intervening), and then spawning a new
variable scope from a counterfactual modification, formalized
as an intervention. This operation effectively resets the 
state of the random variables downstream of the intervention
to their original unbound state. The operation is illustrated 
in Figure~\ref{fig:cf}.

The pseudo-code for the computation of counterfactuals is listed in 
Figure~\ref{alg:counterfactual}. The function \textsc{CF}$(n, m, \delta)$
takes the root~$n$ of the reference probability tree, the root~$m$
of the factual premise tree, and a min-cut~$\delta$ for the
counterfactual event. It returns a new counterfactual probability tree.
The auxiliary function \textsc{Zip}$(\mathcal{A}, \mathcal{B})$ takes
to ordered sets $\mathcal{A} = \{a_n\}_{n=1}^N$ and 
$\mathcal{B} = \{b_n\}_{n=1}^N$ and returns a new
set $\mathcal{C} = \{ (a_n, b_n) \}_{n=1}^N$ of ordered pairs.

\newpage
\section{Discussion}

\paragraph{Previous work.}
As already mentioned in the introduction, the literature
on probability trees is limited. Although trees are common
for representing games (specifically, extensive-form games
\citep{von2007theory, leyton2008essentials}) and 
sequential decision problems \citep{russell2002artificial},
it was Shafer's seminal work~\citep{shafer1996art} 
that first articulated 
a treatment of causality based on probability
trees. Unlike Pearl~\citep{pearl2009causality},
who grounds the semantics of causal relations on the notion of
interventions, Shafer considered causality as a side-effect
entirely subsumed under conditional independence. Thus, to the best
of our knowledge, event-based interventions were only introduced later
independently in~\citep{ortega2015subjectivity}
and in~\citep{gorgen2016differential, gorgen2017algebraic},
the former on systems of nested events (whose algebraic 
closure generates the algebra of the probability space) 
and the latter on an elegant method characterizing
probability trees in terms of interpolating polynomials
(where interventions amount to computing derivatives),
which also establishes the equivalence classes of probability
trees.
The formalization of counterfactuals presented here
is original and differs from the counterfactuals in
\citep{pearl2009causality} in important ways discussed later.

The inability of CBNs (and SCMs) to represent context-specific
independencies is well known: the conditional independencies 
of CBNs are exactly
those that are logically equivalent to a collection of the
form $X_n \indep \{X_1, \ldots, X_{n-1}\} \mid S_n$, where
$X_1, \ldots, X_N$ is an ordering of the vertices in the graph, 
and $S_n$ is a subset of $\{X_1, \ldots, X_{n-1}\}$
\citep{dawid2010beware}. This has led to the suggestion of
using probability trees as an alternative representation
for modeling context-specific independencies
\citep{boutilier2013context, ortega2011bayesian}.
In the context of factor graphs, \citep{minka2009gates} 
introduced a notation (called ``gates'') for modeling
context-specific independencies along with corresponding
inference algorithms. Lastly, since probability trees
may be regarded as a formalization of the computational
traces of probabilistic programs, 
context-specific independencies are modeled naturally 
by probabilistic programming languages~\citep{witty2018causal}.

\paragraph{Computational complexity.}
All of the presented algorithms 
(except \textsc{Neg}, which has~$\mathcal{O}(1)$ time complexity)
follow the same pattern: they recursively descend the
tree once and then backtrack to the root node. The worst-case scenario
occurs when the probability tree is structured like a chain,
in which case the time and space complexity of the algorithms is~$\mathcal{O}(N)$, where~$N$ is the number of nodes in the tree.
However, a shortcoming of our algorithms is that they 
do not exploit the independencies present
in a probability tree, which could dramatically
reduce the computational complexity. More work is required to 
understand how to incorporate those.

\begin{figure}[t]
    \centering
    \includegraphics[width=\textwidth]{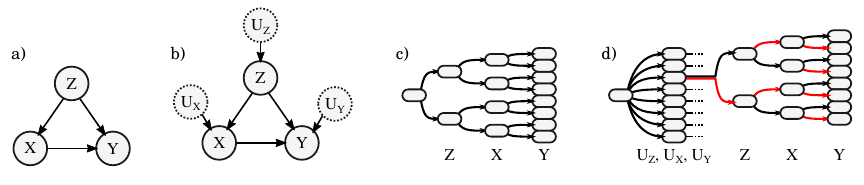}
    \caption{Modeling CBNs versus SCMs as probability trees.
    a) In a CBN, every random variable has an associated probability
    distribution conditioned on the parents, but
    otherwise no further assumptions are made. 
    b) Instead, SCMs assume that the endogenous variables (solid) are 
    deterministic functions of the exogenous random variables (dotted).
    c) The CBN in (a) can be modeled as a four-stage probability
    tree, where each layer binds one of the variables in turn.
    We assumed $X, Y,$ and $Z$ to be binary.
    d) In contrast, the SCM in (b) requires a five-stage
    probability tree, placing all the uncertainty in the
    first transition which binds the triple of exogenous
    variables $(U_Z, U_X, U_Y)$, followed by deterministic
    transitions that bind the endogenous variables $X, Y,$ 
    and $Z$ (the red transitions indicate transitions of
    zero probability).}
    \label{fig:cbn-scm}
\end{figure}

\paragraph{Counterfactuals.}
The algorithm for computing counterfactuals generalizes 
the twin-network construction used in SCMs
\citep{pearl2009causality}. To understand this generalization,
recall that SCMs are stricter than CBNs, in that SCMs impose
constraints on how the uncertainty enters the random experiment
(Figures~\ref{fig:cbn-scm}a,b).
Specifically, the variables of the random experiment---the endogenous 
variables---are deterministic functions of exogenous variables,
which concentrate all the uncertainty. Hence, all the uncertainty of
about the random experiment causally precedes the endogenous
variables. CBNs make no such assumption: in particular,
the uncertainty of any variable could causally depend on
its parents.

This difference matters when estimating counterfactuals.
In an SCM, any knowledge gained about the exogenous variables
transfers to the counterfactual world because it always precedes
any counterfactual intervention on the endogenous variables.
There is no ``endogenous uncertainty''---which would not 
necessarily transfer.

In contrast, in probability trees, there is no 
distinction between endogenous and exogenous variables; 
the uncertainties can originate anywhere along the 
realization of the random experiment in a 
context-specific manner---see e.g.\ how one could model
the difference between CBNs and SCMs in Figures~\ref{fig:cbn-scm}c,d. 
The rule for transferring information
to the counterfactual world is simple: information upstream
of the counterfactual intervention transfers while the one
downstream doesn't. This allows for a fine-grained control
over the assumptions for estimating counterfactuals.

\paragraph{Interpretation.}
Our work presents interpretation-agnostic algorithms for
causal reasoning. For instance, we can impute both
objective and subjective semantics to the probabilities
and causal relations in a probability tree. Regardless,
not every construction yields statements with well-defined
empirical content---e.g.\ see a critique of counterfactual
statements in~\citep{dawid2000causal}. Indeed,
many of the interpretative limitations and caveats that apply
to CBNs extend to probability trees~\citep{dawid2010beware},
especially the reification\footnote{That is, mistaking an abstract idea
for a physical thing.} of the formal elements that are peculiar
to probability trees. More work is required in order
to understand these.

It must be noted however, that a subjective interpretation of
probability trees puts causal discovery on a firm
formal ground. Questions such as ``does $X$ cause $Y$ or
$Y$ cause $X$?'' can both be articulated (through conditional
causal relations) and answered (at least in a Bayesian sense,
through observing the effects of interventions)
without introducing extraneous formal machinery 
\citep{ortega2011bayesian}. Unlike in CBNs, there is no need
to reason over a collection of models---a single probability
tree suffices.

\paragraph{Features and Limitations.}
The presented algorithms assume that probability trees are finite.
If the random variables are well-formed (Section~\ref{sec:mincut}),
then the algorithms extend without modification to trees
with countably many nodes, because the computations of min-cuts
and the causal reasoning operations always terminate due to their
recursive nature. Extensions to trees with uncountably many nodes
require fundamentally different techniques, but they are
strictly more general than CBNs in their expressive power.

The focus on events rather than random variables
has led us to rethink and generalize the causal 
reasoning operations. This is especially salient
in the algorithms for computing interventions and 
counterfactuals. The intervention algorithm presented here
allows answering reverse-engineering questions like ``what 
are the necessary manipulations in order to bring about 
an event $A$?''. Moreover, though the resulting probabilities
coincide when applicable, the computation of counterfactuals in
probability trees differs significantly from that in CBNs
using e.g.\ the twin network construction~\citep{pearl2009causality}.

Having said that, the building blocks for constructing
events are still limited.
In particular, they are limited to propositional logic
and one causal relation, namely \emph{precedence}
(Section~\ref{sec:mincut-causal}). We envision
the addition of more causal relations, and the extension
to first-order logic, to enrich the language of events. 
In particular, one could also explore links 
to~\emph{temporal logic}~\citep{ohrstrom2007temporal}
and \emph{probabilistic programming}~\citep{van2018introduction}.

A system of transformations akin to the \emph{do-calculus}
is entirely absent and left for future work. Such a
system is necessary for addressing questions about
\emph{identification}, i.e.\ whether a causal effect
can be estimated from observation alone~\citep{pearl2009causality}.
More generally, the precise algebraic properties of the causal
operations, such as when conditions and interventions commute,
are currently unknown.

Finally, the decision to place uniform conditional probabilities over
events with probability zero is arbitrary and only made
for the sake of simplicity and concreteness. To encode other
versions of the conditional probabilities, one could ban
zero-probability transitions and replace them with ``vanishingly
small'' quantities (for instance, polynomials 
in~$\epsilon \approx 0$).

\section*{Acknowledgments}
We thank Tom Everitt, Ramana Kumar for their comments and Silvia Chiappa for proof-reading. We also thank Judea Pearl for providing an example which led us to discover an error in our algorithm for computing counterfactuals.

\bibliographystyle{unsrt}
\bibliography{bibliography}

\end{document}